\documentclass{article}
\usepackage{ijcai16}

\usepackage{times}

\usepackage{epsfig}
\usepackage{graphicx}
\usepackage{subfigure}
\usepackage{amsmath}
\usepackage{amssymb}

\usepackage{mathrsfs}
\usepackage{algorithm}
\usepackage{algorithmic}
\usepackage{array}
\usepackage{booktabs}
\usepackage{multirow}
\usepackage[super]{nth}

\newcommand{\tabincell}[2]{\begin{tabular}{@{}#1@{}}#2\end{tabular}}
\title{Geometric Scene Parsing with Hierarchical LSTM }
\author{Zhanglin Peng$^{1}$, Ruimao Zhang$^{1}$, Xiaodan Liang$^{1}$, Xiaobai Liu$^{2}$, Liang Lin$^{1}$\thanks{Corresponding author is Liang Lin (Email: linliang@ieee.org).This work was supported in part by Special Program for Applied Research on Super Computation of the NSFC-Guangdong Joint Fund (the second phase), in part by Guangdong Natural Science Foundation under Grant S2013050014548, in part by the Fundamental Research Funds for the Central Universities.} \\
$^{1}$Sun Yat-sen University, Guangzhou, China \\
$^{2}$San Diego State University, U.S.
}

\begin{document}

\maketitle

\begin{abstract}
This paper addresses the problem of geometric scene parsing, i.e. simultaneously labeling geometric surfaces (e.g. sky, ground and vertical plane) and determining the interaction relations (e.g. layering, supporting, siding and affinity) between main regions. This problem is more challenging than the traditional semantic scene labeling, as recovering geometric structures necessarily requires the rich and diverse contextual information. To achieve these goals, we propose a novel recurrent neural network model, named Hierarchical Long Short-Term Memory (H-LSTM). It contains two coupled sub-networks: the Pixel LSTM (P-LSTM) and the Multi-scale Super-pixel LSTM (MS-LSTM) for handling the surface labeling and relation prediction, respectively. The two sub-networks provide complementary information to each other to exploit hierarchical scene contexts, and they are jointly optimized for boosting the performance. Our extensive experiments show that our model is capable of parsing scene geometric structures and outperforming several state-of-the-art methods by large margins. In addition, we show promising 3D reconstruction results from the still images based on the geometric
parsing.

\end{abstract}

\vspace{-0.8em}
\section{Introduction}

%Geometric scene parsing refers to simultaneously assign a geometric surface label (e.g. sky, ground, vertical plane) to each pixel and geometric relations (e.g. layering, supporting, siding and affinity) to two adjacent geometric regions, as illustrated in Figure~\ref{fig:SceneParsing}. Based on the predicted geometric surface regions and relation results, the 3D scene can be conveniently reconstructed from a still image. This fundamental task enables the intelligence system to understand the geometry structure of the scene in details. It can help many different levels' visual applications, such as single-view 3D reconstruction~\cite{Conf:SingleView}, surface normal estimation~\cite{Conf:SurfaceNormal}, indoor navigation~\cite{Conf:IndoorNavigation} and Robots~\cite{Conf:Robots}. Although a series of works~\cite{Conf:RecurrentNNforlabeling}\cite{Conf:FCN}\cite{Conf:ImageToPixelLabel}  have been proposed to deal with the semantic segmentation/labelling task, geometric scene parsing remains largely under explored due to the following challenges:

%Despite considerable efforts been made, 3D Reconstruction from a single-view image is still an open problem due to the reconstruction information recovered from single image is limited.

\begin{figure}[t]
\centering
\includegraphics[width= 2.8 in]{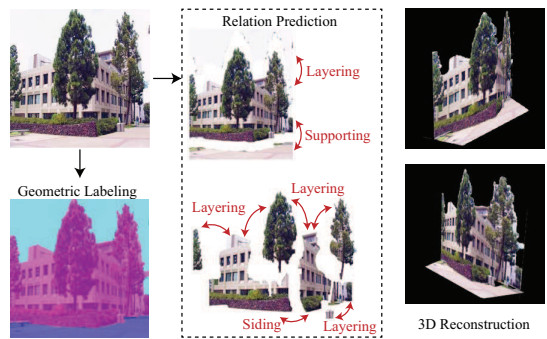}\vspace{-1em}
   \caption{An illustration of our geometric scene parsing. Our task aims to predict the pixel-wise geometric surface labeling (first column) and the interaction relations between main regions (second column). Then the parsing result is applied to reconstruct a 3D model (third column).}\vspace{-1em}
\label{fig:SceneParsing}
\end{figure}

Humans can naturally sense the geometric structures of a scene by a single glance, while developing such a system remains to be quite challenging in several intelligent applications such as robotics \cite{Conf:Robots} and automatic navigation \cite{Conf:IndoorNavigation} . In this work, we investigate a novel learning-based approach for geometric scene parsing, which is capable of simultaneously labeling geometric surfaces (e.g. sky, ground and vertical) and determines the interaction relations (e.g. layering, supporting, siding and affinity \cite{Conf:SingleView}) between main regions, and further demonstrate its effectiveness in 3D reconstruction from a single scene image. An example generated by our approach is presented in Figure~\ref{fig:SceneParsing}. In the literature of scene understanding, most of the efforts are dedicated for pixel-wise semantic labeling / segmentation \cite{Conf:FCN}\cite{Conf:ImageToPixelLabel}. Although impressive progresses have been made, especially by the deep neural networks, these methods may have limitations on handling the geometric scene parsing due to the following challenges.

%Geometric scene parsing, which refers to simultaneously assign a geometric surface label (e.g. sky, ground, vertical plane) to each pixel and geometric relations (e.g. layering, supporting, siding and affinity) to a couple of adjacent geometric regions, provides a wealth of information for scene image understanding and enables the intelligence system to know the geometry structure in details. It helps many different levels' visual applications, such as single-view 3D reconstruction~\cite{Conf:SingleView}, surface normal estimation~\cite{Conf:SurfaceNormal}, indoor navigation~\cite{Conf:IndoorNavigation} and Robots~\cite{Conf:Robots}. Among them, 3D reconstruction from a single-view image is still an open problem due to the reconstruction information recovered from single image is limited. As illustrated in Figure~\ref{fig:SceneParsing}, our purpose is to parse the single image into the geometric surfaces and relations, and further conveniently reconstruct 3D scene according to these structured information.  In literature, although a series of works~\cite{Conf:RecurrentNNforlabeling}\cite{Conf:FCN}\cite{Conf:ImageToPixelLabel}  have been proposed to deal with the semantic segmentation/labelling task, geometric scene parsing remains largely under explored due to the following challenges:

\begin{figure*}
\centering
\includegraphics[width= 6 in]{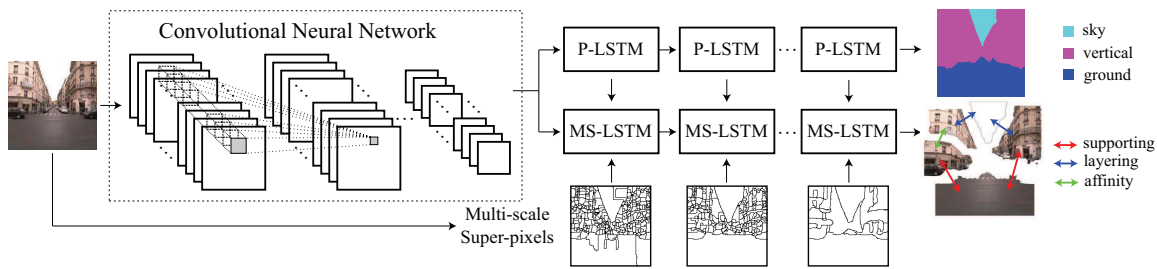}\vspace{-1em}
\caption{The proposed recurrent framework for geometric scene parsing. Each still image is first fed into several convolutional layers. Then these feature maps are passed into the the stacked Pixel LSTM (P-LSTM) layers and Multi-scale Super-pixel LSTM( MS-LSTM) to generate the geometric surface labeling of each pixel and interaction relations between regions, respectively. }\vspace{-1em}
\label{fig:framewrok}
\end{figure*}

\begin{itemize}

\item The geometric regions in a scene often have diverse appearances and spatial configurations, e.g. the vertical plane may include trees and buildings of different looks. Labeling these regions generally requires fully exploiting image cues from different aspects ranging from local to global.

\item In addition to region labeling, discovering the interaction relations between the main regions is crucial for recovering the scene structure in depth. The main difficulties for the relation prediction lie in the ambiguity of multi-scale region grouping and the fusion of hierarchical contextual information.

\end{itemize}

To address these above issues, we develop a novel \textbf{Hierarchical LSTM} (H-LSTM) recurrent network that simultaneously parses a still image into a series of geometric regions and predicts the interaction relations among these regions. The parsing results can be directly used to reconstruct the 3D structure from a single image. As shown in Figure~\ref{fig:framewrok}, the proposed model collaboratively integrates the Pixel LSTM (P-LSTM) \cite{Conf:LG-LSTM} and Multi-scale Super-pixel LSTM (MS-LSTM) sub-networks into a unified framework. First, the P-LSTM sub-network produces the geometric surface regions, where local contextual information from neighboring positions is imposed on each pixel to better exploit the spatial dependencies. Second, the Multi-scale Super-pixel LSTM (MS-LSTM) sub-network generates the interaction relations for all adjacent surface regions based on the multi-scale super-pixel representations. Benefiting from the diverse levels of information captured by hierarchical representations (i.e. pixels and multi-scale super-pixels), the proposed H-LSTM can jointly optimize the two tasks based on the hierarchical information, where different levels of contexts are captured for better reasoning in local area. Based on the shared basic convolutional layers, the parameters in P-LSTM and MS-LSTM sub-networks are jointly updated during the back-propagation. Therefore, the pixel-wise geometric surface prediction and the super-pixel-wise relation categorization can mutually benefit from each other.

The proposed H-LSTM is primarily inspired by the success of  Long Short-Term Memory Networks (LSTM) \cite{Conf:MultidimensionalRNN} \cite{Arxiv:GridLSTM} \cite{Conf:LG-LSTM} on the effective incorporation of long and short rang dependencies  from the whole image. Different from previous LSTM structure \cite{Conf:2DLSTM} \cite{Conf:LSTM-Labeling} \cite{Conf:LG-LSTM} that simply operates on each pixel,  our H-LSTM exploits hierarchical information dependencies from different levels of units such as pixels and multi-scale super-pixels. The hidden cells are treated as the enhanced features and the memory cells can recurrently remember all previous contextual interactions for different levels of representations from different layers.

%and the accuracy of surface labelling is promoted effectively.

Since the geometric surface labeling needs the fine prediction results while the relation prediction cares more about the coarse semantic layouts, we thus resort to the specialized P-LSTM and MS-LSTM to separately address these two tasks. In terms of \textbf{geometric surface labeling}, the P-LSTM is used to incorporate the information from neighboring pixels to guide the local prediction of each pixel, where the local contextual information can be selectively remembered and then guide the feature extraction in the later layer. In terms of \textbf{ interaction relation prediction}, the MS-LSTM effectively reduces the information redundancy by the natural smoothed regions and different levels of information can be hierarchically used to extract interaction relations in different layers. Particularly, in each MS-LSTM layer, the super-pixel map with a specific scale is used to extract the smoothed feature representation. Then, the features of adjacent super-pixels are fed into the LSTM units to exploit the spatial dependencies. The super-pixel map with larger scale is used in the deep layer to extract the higher-level contextual dependencies. After passing through all of the hierarchical MS-LSTM layers, the final interaction relation prediction can be obtained by the final relation classifier based on the enhanced features benefiting from the hierarchical LSTM units.

%Note that if the siding relation exists in the coupled regions in the last layer, we will merge them into the one.

% Since the geometric surface labeling needs the fine prediction results while the interaction relation predictions care more about the coarse semantic layouts, we thus resort to the specialized P-LSTM and MS-LSTM to separately address these two tasks.

This paper makes the following three contributions. (1) A novel recurrent neural network model is proposed for geometric scene parsing, which jointly optimizes the geometric surface labeling and relation prediction. (2) Hierarchically modeling image contexts with LSTM units over super-pixels is original to the literature, which can be extended to similar tasks such as human parsing. (3) Extensive experiments on three public benchmark demonstrate the superiority of our H-LSTM model over other state-of-the-art geometric surface labeling approaches. Moreover, we show promising 3D reconstruction results from the still images based on the geometric parsing.

\vspace{-0.5em}
\section{Related Work}

\textbf{Semantic Scene Labeling.} Most of the existing works focused on the semantic region labeling problem \cite{Conf:CRF1} \cite{Conf:Recursive_Socher} \cite{Conf:FCN}, while the critical interaction relation prediction is often overlooked. Based on the hand-crafted features and models, the CRF inference \cite{Conf:CRF2} \cite{Conf:CRF1} refines the labeling results by considering the label agreement between similar pixels. The fully convolutional network (FCN) \cite{Conf:FCN} and its expansion \cite{Conf:DeepLab} have achieved great success on the semantic labeling. \cite{Conf:CNN-MRF} incorporates the markov random field (MRF) into deep networks for pixel-level labeling. Most recently, the multi-dimensional LSTM \cite{Conf:LSTM-Labeling} has also been employed to capture the local spatial dependencies. However, our H-LSTM differs from these works in that we train a unified network to collaboratively address the geometric region labeling and relation prediction. The novel P-LSTM and MS-LSTM can effectively capture the long-range spatial dependencies benefiting from the hierarchical feature representation on the pixels and multi-scale super-pixels.

%These methods are able to produce the sharp boundaries and revise the incorrect pixel-level label predictions.

\textbf{Single View 3D Reconstruction.} The 3D reconstruction from the singe view image is an under explored task and only a few researches have made some efforts on this task. Mobahi et al. \cite{Conf:Holistic3D} reconstructed the urban structures from the single view by transforming invariant low-rank textures. Without the explicit assumptions about the structure of the scene, Saxena et al. \cite{DBLP:Make3D} trained the MRF model to discover the  depth cues as well as the relationships between different parts of the image in a fully supervised manner. An attribute grammar model \cite{Conf:SingleView} regarded super-pixels as its terminal nodes and applied five production rules to generate the scene into a hierarchical parse graph. Differed from the previous methods, the proposed H-LSTM  predicts the layout segmentation and the spatial arrangement with a unified network architecture, and thus can reconstruct the 3D scene from a still image directly.

\vspace{-0.3em}
\section{Hierarchical LSTM}
\label{sec:hlstm}

% Overview 感觉和前边有点重复，需要确认下到底要不要

\textbf{Overview.} The geometric scene parsing aims to generate the pixel-wise geometric surface labeling and relation prediction for each image. As illustrated in Figure~\ref{fig:framewrok}, the input image is first passed through a stack of convolutional and pooling layers to generate a set of convolutional feature maps. Then the P-LSTM and MS-LSTM take these feature maps as inputs in a share mode, and their outputs are the pixel-wise geometric surface labeling and interaction relations between adjacent regions respectively.

%The traditional LSTM~\cite{DBLP:Tr-LSTM} is introduced to memorize the long-period interdependencies in sequential data. Such temporal dependency is also elegantly converted to the spatial domain~\cite{Conf:LSTM-Labeling}\cite{Conf:LG-LSTM}.

\noindent
\textbf{Notations.}  Each LSTM \cite{LSTM} unit in $i$-th layer receives the input $\mathbf{x}_i$ from the previous state, and determines the current state which is comprised of the hidden cells $\mathbf{h}_{i+1}\in R^d$ and the memory cells $\mathbf{m}_{i+1}\in R^d$, where $d$ is the dimension of the network output. Similar to the work in \cite{Conf:Speech-RNN}, we apply $g^u$,$g^f$,$g^v$,$g^o$ to indicate the input, forget, memory and output gate respectively. Define $W^u$,$W^f$,$W^v$,$W^o$ as the corresponding recurrent gate weights. Thus the hidden and memory cells for the current state can be calculated by,
\begin{equation}\label{Eq:lstm}
\begin{split}
& g^u = \phi(W^u*\mathbf{H}_i)  \\
& g^f = \phi(W^f*\mathbf{H}_i)  \\
& g^o = \phi(W^o*\mathbf{H}_i)    \\
& g^v = \tanh(W^v*\mathbf{H}_i)    \\
& \mathbf{m}_{i+1} = g^f\odot \mathbf{m}_i + g^u\odot g^v \\
& \mathbf{h}_{i+1} = \tanh(g^o\odot \mathbf{m}_i)
\end{split}
\end{equation}
where $\mathbf{H}_i$ denotes the concatenation of input $\mathbf{x}_i$ and previous state $\mathbf{h}_i$. $\phi$ is a sigmoid function with the form $\phi(t) = 1/ (1+e^{-t})$, and $\odot$ indicates the element-wise product. Following \cite{Arxiv:GridLSTM}, we can simplify the expression Eqn.(\ref{Eq:lstm}) as,
\begin{equation}\label{Eq:SimpleLSTM}
(\mathbf{m}_{i+1},\mathbf{h}_{i+1}) = \textbf{LSTM}(\mathbf{H}_i,\mathbf{m}_i,W)
\end{equation}
where $W$ is the concatenation of four different kinds of recurrent gate weights.

\vspace{-0.5em}
\subsection{P-LSTM for Geometric Surface Labeling}

Following \cite{Conf:LG-LSTM}, we use the P-LSTM to propagate the local information to each position and further discover the short-distance contextual interactions in pixel level. For the feature representation of each position $j$, we extract $N=8$ spatial hidden cells from $N$ local neighbor pixels and one depth hidden cells from previous layer. Note that the ``depth" in a special position indicates the features produced by the hidden cells at that position in the previous layer. Let $\{\mathbf{h}_{j,i,n}^s\}_{n=1}^N$ indicate the set of hidden cells from neighboring positions to pixel $j$, which are calculated by the $N$ spatial LSTMs updated in $i$-th layer. And $\mathbf{h}_{j,i}^t$ denotes the hidden cells computed by the $i$-th layer depth LSTM on the pixel $j$. Then the input states of pixel $j$ for the $(i+1)$-th layer LSTM can be expressed by,
\begin{equation}\label{Eq:H_ij}
\mathbf{H}_{j,i} = [~\mathbf{h}_{j,i,1}^s ~~~ \mathbf{h}_{j,i,2}^s ~~... ~~\mathbf{h}_{j,i,n}^s ~~~ \mathbf{h}_{j,i}^t ~]^T
\end{equation}
where $\mathbf{H}_{j,i} \in R^{(N+1)\times d}$. By the same token, let $\{\mathbf{m}_{j,i,n}^s\}_{n=1}^N$ be the memory cells for all $N$ spatial dimensions of pixel $j$ in the $i$-th layer and $\mathbf{m}_{j,i}^t$ be memory cell for the depth dimension. Then the hidden cells and memory cells of each position $j$ in the $(i+1)$-th layer for all $N+1$ dimensions are calculated as,
\begin{equation}\label{Eq:update}
\begin{split}
 (\mathbf{m}_{j,i+1,n}^s~,~\widetilde{\mathbf{h}}_{j,i+1,n}^s)& = \textbf{LSTM}(\mathbf{H}_{j,i}~,~\mathbf{m}_{j,i,n}^s~,~W_i^s) \\
 n& \in \{1,2,...,N\}; \\
 (\mathbf{m}_{j,i+1}^t~,~\mathbf{h}_{j,i+1}^t)& = \textbf{LSTM}(\mathbf{H}_{j,i}~,~\mathbf{m}_{j,i}^t~,~W_i^t)
\end{split}
\end{equation}
where $W_i^s$ and $W_i^t$ indicate the weights for spatial and depth dimension in the $i$-th layer, respectively. Note that $\widetilde{h}_{j,i+1,n}^s$ should be distinguished from ${h}_{j,i+1,n}^s$ by the directions of information propagation. $\widetilde{h}_{j,i+1,n}^s$ represents the hidden cells position $j$ to its $n$-th neighbor, which is used to generate the input hidden cells of $n$-th neighbor position for the next layer. In contrast, ${h}_{j,i+1,n}^s$ is the neighbor hidden cells fed into Eqn.(\ref{Eq:H_ij}) to calculate the input state of pixel $j$.

In particular, the P-LSTM sub-network is built upon the modified VGG-16 model \cite{Conf:VGG-16}. We remove the last two fully-connected layers in VGG-16, and replace with two fully-convolutional layers to obtain the convolutional feature maps for the input image. Then the convolutional feature maps are fed into the transition layer \cite{Conf:LG-LSTM} to produce hidden cells and memory cells of each position in advance, and make sure the number of the input states for the first P-LSTM layer is equal to that of following P-LSTM layer. Then the hidden cells and memory cells are passed through five stacked P-LSTM layers. By this way, the receptive field of each position can be considerably increased to sense a much larger contextual region. Note that the intermediate hidden cells generated by P-LSTM layer are also taken as the input to the corresponding Super-pixel LSTM layer for relation prediction. Please check more details of this part in Sec.~\ref{Sec:Super-pixel LSTM}. At last, several $1\times 1$ feed-forward convolutional filters are applied to generate confidence maps for each geometric surface. The final label of each pixel is returned by a softmax classifier with the form,
\begin{equation}\label{Eq:LabelPrediction}
y_{j} = \text{softmax} ( \mathcal{F}(~\mathbf{h}_{j};W_{label}))
\end{equation}
where $y_{j}$ is the predicted geometric surface probability of the $j$-th pixel, and $W_{label}$ denotes the network parameter. $\mathcal{F}(\cdot)$ is a transformation function.

\vspace{-0.5em}
\subsection{MS-LSTM for Interaction Relation Prediction}
\label{Sec:Super-pixel LSTM}

\begin{figure}[t]
\centering
\includegraphics[width= 3.3 in]{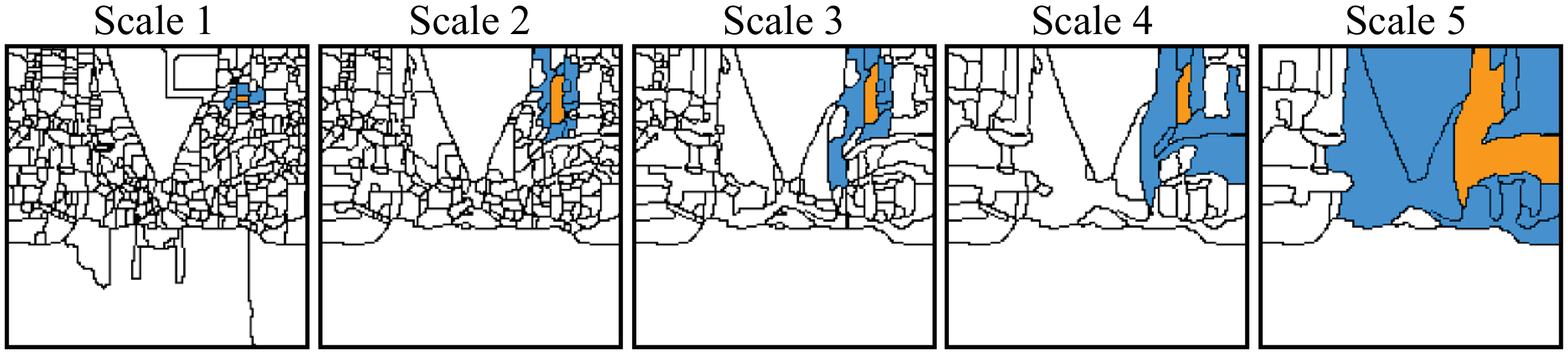}\vspace{-1em}
   \caption{An illustration of super-pixel maps with different scales. In each scale, the orange super-pixel is the one under the current operation, and the blue ones are adjacent super-pixels, which propagate the neighboring information to the orange one. More contextual information can be captured by the larger-scale super-pixels.}\vspace{-1em}
\label{fig:p-LSTM}
\end{figure}

The Multi-scale Super-pixel LSTM (MS-LSTM) is used to explore high-level interaction relation between pair-wise super-pixels, and predict the functional boundaries between geometric surfaces. The hidden cells of $j$-th position in $i$-th MS-LSTM layer are the concatenation of hidden cells $\mathbf{h}_{j,i}^t \in R^d$  from previous layer (same as the depth dimension in P-LSTM) and $\mathbf{h}_{j,i}^r \in R^d$  from the corresponding P-LSTM layer. For simplicity, we rewrite the enhanced hidden cells as $\mathbf{\hbar}_{j,i}=[~ \mathbf{h}_{j,i}^t ~,~ \mathbf{h}_{j,i}^r ~]$. In each MS-LSTM layer, an over-segmentation algorithm \cite{Conf:SuperPixelGeneration} is employed to produce the super-pixel map $S^i$ with a specific scale $c_i$. To obtain the compact feature representation for each super-pixel, we use \textit{Log-Sum-Exp}(LSE) \cite{Book:LSE}, a convex approximation of the \textit{max} function to fuse the hidden cells of pixels in the same super-pixel,
\begin{equation}\label{Eq:fusion}
\mathbf{h}_{\Lambda,i} = \frac{1}{\pi}\log \left[\frac{1}{Q_\Lambda}\sum_{j \in \Lambda} \exp (\pi \mathbf{\hbar}_{j,i}) \right]
\end{equation}
where $\mathbf{h}_{\Lambda,i} \in R^{2d}$ denotes the hidden cells of the super-pixel $\Lambda$ in the $i$-th super-pixel layer, $\mathbf{\hbar}_{j,i}$ denotes the enhance hidden cells of the $j$-th position, $Q_\Lambda$ is the total number of pixels in $\Lambda$ , and $\pi$ is a hyper-parameter to control smoothness. With higher value of $\pi$, the function tends to preserve the max value for each dimension in the hidden cells, while with lower value the function behaves like an averaging function.

Similar to the Eqn.(\ref{Eq:H_ij}), let $\{\mathbf{h}_{\Lambda,i,k}\}_{k=1}^{K_{\Lambda}}$ indicate the set of hidden cells from $K_{\Lambda}$ adjacent super-pixels of $\Lambda$. Then the input states of super-pixel $\Lambda$ for the $(i+1)$-th MS-LSTM layer can be computed by,
\begin{equation}\label{Eq:H_superpixel}
\mathbf{H}_{\Lambda,i} = [~ \frac{1}{K_{\Lambda}}\sum_k \mathbf{h}_{\Lambda,i,k} ~~~ \mathbf{h}_{\Lambda,i} ~]^T
\end{equation}
where $\mathbf{H}_{\Lambda,i} \in R^{4d}$. The hidden cells and memory cells of super-pixel $\Lambda$ in the $(i+1)$-th layer can be calculated by,
\begin{equation}\label{Eq:Update2}
(\mathbf{m}_{\Lambda,i+1}~,~\mathbf{h}_{\Lambda,i+1}) = \textbf{LSTM}(\mathbf{H}_{\Lambda,i}~,~\mathbf{m}_{\Lambda,i}~,~W_i^{'})
\end{equation}
where $W_i^{'}$ denotes the concatenation gate weights of $i$-th MS-LSTM layer. $\mathbf{m}_{\Lambda,i}$ is the average value of the memory cells of each position in super-pixel $\Lambda$. Note that the dimension of   $\mathbf{h}_{\Lambda.i+1}$ in Eqn.(\ref{Eq:Update2}) is $d$, which is equal to the output hidden cells from the P-LSTM. In the $(i+1)$-th layer, the values of $\mathbf{h}_{\Lambda,i+1}$ and $\mathbf{m}_{\Lambda,i+1}$  can be directly assigned to the hidden cells and memory cells of each position in super-pixel $\Lambda$. Then the new hidden states can be accordingly learned by applying MS-LSTM layer on the super-pixel map with larger scale.

In particular, the MS-LSTM layers share the convolutional feature maps with the P-LSTM. In total, five stacked MS-LSTM layers are applied to extract hierarchical feature representations with different scales of contextual dependencies. Therefore, five super-pixel maps with different scales (i.e. 16, 32, 48, 64 and 128) are extract by the over-segmentation algorithm \cite{Conf:SuperPixelGeneration}. Note that the scale in here refers to the average number of pixels in each super-pixel. Thus these multi-scale super-pixel maps are employed by different MS-LSTM layers, and the hidden cells for each layer are enhanced by the output of the corresponding P-LSTM layer. After passing though these hierarchical MS-LSTM layers, the local inference of each super-pixel can be influenced by different degrees of context, which enables the model simultaneously taking the local semantic information into account. Finally, the interaction relation prediction of adjacent super-pixels is optimized as,
\begin{equation}\label{Eq:RelationPrediction}
z_{\{\Lambda,\Lambda'\}} = \text{softmax} (\mathcal{F}([~\mathbf{h}_{\Lambda}~\mathbf{h}_{\Lambda'}];W_{relation}^{'}))
\end{equation}
where $z_{\{\Lambda,\Lambda'\}}$ is the predicted relation probability vector between super-pixel $\Lambda$ and $\Lambda'$, and $W_{relation}^{'}$ denotes the network parameters. $\mathcal{F}(\cdot)$ is a transformation function.

%Both the merging and dividing operations are effectively used to refine the final geometric labeling results.
\vspace{-0.5em}
\subsection{Model Optimization}

The total loss of H-LSTM is the sum of losses of two tasks: geometric surface labeling loss $\mathcal{J}_C$ by P-LSTM and relation prediction loss  $\mathcal{J}_R$ by MS-LSTM. Given $U$ training images with $\{(I_1,\widehat{Y}_1,\widehat{Z}_1),...,(I_U,\widehat{Y}_U, \widehat{Z}_U)\}$, where $\widehat{Y}$ indicates the groundtruth geometric surfaces for all pixels for image $I$,and $\widehat{Z}$ denotes the groundtruth relation labels for all of adjacent super-pixel pairs in different scales. The overall loss function is as follows,
\begin{equation}\label{Eq:TotalLoss}
\mathcal{J}(W) = \frac{1}{U} \sum_{i=1}^U (  \mathcal{J}_C(W_P;I_i,\widehat{Y}_i) + \mathcal{J}_R(W_S;I_i,\widehat{Z}_i)    )
\end{equation}
where $W_P$ and $W_S$ indicate the parameters of P-LSTM and MS-LSTM, respectively, and $W$ denotes all of the parameters with the form $W=\{W_P,W_S,W_{CNN}\}$. $W_{CNN}$ is the parameters of Convolution Neural Network. We apply the back propagation algorithm to update all the parameters. $ \mathcal{J}_C(\cdot)$ is the standard pixel-wise cross-entropy loss. $ \mathcal{J}_R(\cdot)$ is the cross-entropy loss for all super-pixels under all scales. Each MS-LSTM layer with a specific scale of the super-pixel map can output the final interaction relation prediction. Note that $ \mathcal{J}_R(\cdot)$ is the sum of losses after all MS-LSTM layers.

\vspace{-0.3em}
\section{Application to 3D Reconstruction}

In this work, we apply our geometric scene parsing results for single-view 3D reconstruction. The predicted geometric surfaces and their relations are used to "cut and fold" the image into a pop-up model \cite{DBLP:Reconstruction}. This process contains two main steps: (1) restoring the 3D spatial structure based on the interaction relations between adjacent super-pixels, (2) constructing the positions of the specific planes using projective geometry and texture mapping from the labelled image onto the planes. In practice, we first find the ground-vertical boundary according to the predicted supporting relations and estimate the horizon position as the benchmark of 3D structure. Then the algorithm uses the different kinds of predicted relations to generate the polylines and folds the space along these polylines. The algorithm also cuts the ground-sky and vertical-sky boundaries according to the layering relations. At last, the geometric surface is projected onto the above 3D structures to reconstruct the 3D model.

 %Our reconstruction method dependents on the work in~\cite{DBLP:Reconstruction}, but the different is that our approach can reconstruct the 3D model from the output of proposed H-LSTM (geometric surfaces and relations) immediately without estimating the polyline based on the geometric labelling and a set of assumptions.

%Finally, the geometric relations between the semantic surface regions are by combining the prediction results of all super-pixel pairs. The adjacent super-pixels with the affinity (or siding) relation are merged into a larger region, while the adjacent super-pixels with the layering (or supporting) relation are divided into different geometric surface.

\vspace{-0.3em}
\section{Experiment}

%In this section, we evaluate the performance of the proposed method. We first describe the experimental setup, and then report the results on the geometric labeling and the correlation prediction, respectively, and finally show some results of single view 3D reconstruction.

\subsection{Experiment Settings}

\textbf{Datasets.} We validate the effectiveness of  the proposed H-LSTM on three public datasets, including \textbf{SIFT-Flow} dataset \cite{DBLP:SiftFlow}, \textbf{LM+SUN} dataset \cite{DBLP:LM+SUN} and \textbf{Geometric Context} dataset \cite{hoiem2007recovering}. The \textbf{SIFT-Flow} consists of 2,488 training images and 200 testing images. The \textbf{LM+SUN} contains 45,676 images (21,182 indoor images and 24,494 outdoor images), which is derived by mixing part of SUN dataset \cite{Conf:SunDataset} and LabelMe dataset \cite{DBLP:LabelMe}. Following \cite{DBLP:LM+SUN}, we apply 45,176 images as training data and 500 images as test ones. For these two datasets, three geometric surface classes (i.e. \textit{sky}, \textit{ground} and \textit{vertical}) are considered for the evaluation. The \textbf{Geometric Context} dataset includes 300 outdoor images, where 50 images are used for training and the rest for testing as \cite{Conf:SingleView}. Except for the  three main geometric surface classes as used  in the previous two datasets, \textbf{Geometric Context} dataset also labels the five subclasses: \textit{left}, \textit{center}, \textit{right}, \textit{porous}, and \textit{solid} for \textit{vertical} class. For all of three datasets, four interaction relation labels (i.e. \textit{layering}, \textit{supporting}, \textit{siding} and \textit{affinity}) are defined and evaluated in our experiments.

\noindent
\textbf{Evaluation Metrics.} Following \cite{Conf:FCN}, we use the pixel accuracy and mean accuracy metrics as the standard evaluation criteria for the geometric surface labeling. The pixel accuracy assesses the classification accuracy of pixels over the entire dataset while the mean accuracy calculates the mean accuracy for all categories. To evaluate the  performance of relation prediction, the average precision metric is adopted.

\noindent
\textbf{Implementation Details.} In our experiment, we keep the original size $256 \times 256$ of the input image for the SIFT-Flow dataset. The scale of input image is fixed as $321 \times 321$ for LM+SUN and Geometric Context datasets. All the experiments are carried out on a PC with NVIDIA Tesla K40 GPU, Intel Core i7-3960X 3.30GHZ CPU and 12 GB memory. During the training phase, the learning rates of transition layer, P-LSTM layers and MS-LSTM layers are initialized as $0.001$ and that of pre-training CNN model is initialized as $0.0001$. The dimension of hidden cells and memory cells, which is corresponding to the symbol $d$ in Sec.~\ref{sec:hlstm}, is set as 64 in both P-LSTM and MS-LSTM.

\vspace{-0.5em}
\subsection{Performance Comparisons}

\begin{figure}[t]
\centering
\includegraphics[width= 2.9 in]{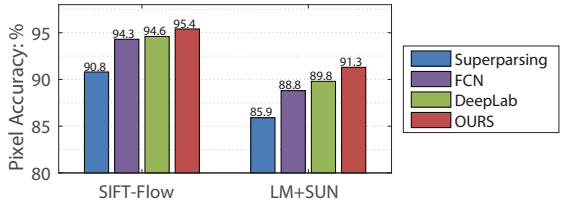}\vspace{-1.2em}
   \caption{Geometric surface labeling results (Pixel-wise Accuracy) on SIFT-Flow and LM+SUN datasets.}\vspace{-0.5em}
\label{fig:PixeAccuracy}
\end{figure}

\textbf{Geometric Surface Labeling.}  We compare the proposed H-LSTM with three recent state-of-the-art approaches, including Superparsing \cite{DBLP:LM+SUN}, FCN \cite{Conf:FCN} and DeepLab~\cite{Conf:DeepLab} on the SIFT-Flow and LM+SUN datasets. Figure \ref{fig:PixeAccuracy} gives the the comparison results on the pixel accuracy. Table \ref{table:SIFT-Flow} and Table \ref{table:LM+SUN} show the performance of our H-LSTM and comparisons with three state-of-the-art methods on the per-class accuracy. It can be observed that the proposed H-LSTM can significantly outperform three baselines in terms of both metrics. For the Geometric Context dataset, the model is fine-tuned based on the trained model on LM+SUN  due to the small size of training data. We compare our results with those reported in \cite{hoiem2008closing}, \cite{DBLP:LM+SUN} and \cite{Conf:SingleView}. Table \ref{table:GeometricContext} reports the pixel accuracy on three main classes and five subclasses. Our H-LSTM can outperform the three  baselines over $3.8\%$ and $2.8\%$ when evaluating on three main classes and five subclasses, respectively.  This superior performance achieved by H-LSTM on three public datasets demonstrates that incorporating the coupled P-LSTM and MS-LSTM in a unified network is very effective in capturing the complex contextual patterns within images that are critical to exploit the diverse surface structures.

\begin{table}\small
\renewcommand{\arraystretch}{1.1}
\addtolength{\tabcolsep}{-1pt}
\begin{center}
\begin{tabular}{c|cccc}
\hline
 Method             & Sky & Ground & Vertical & Mean Acc.  \\
\hline
Superparsing  & - & - & -  &     89.2        \\
FCN         & 96.4 & 93.1 & 91.8  & 93.8         \\
DeepLab      & 96.1 & 93.8 & 93.4 & 94.4           \\
\hline
Ours         & \textbf{96.4} & \textbf{95.1}  & \textbf{93.1}  & \textbf{94.9}            \\
\hline
\end{tabular}\vspace{-0.5em}
\caption{Comparison of geometric surface labeling performance with three state-of-the-art methods on SIFT-Flow dataset.}\vspace{-0.5em}
\label{table:SIFT-Flow}
\end{center}
\end{table}

\begin{table}\small
\renewcommand{\arraystretch}{1.1}
\addtolength{\tabcolsep}{-1pt}
\begin{center}
\begin{tabular}{c|cccc}
\hline
 Method             & Sky & Ground & Vertical & Mean Acc.  \\
\hline
Superparsing    & - & - & - &  86.8      \\
FCN             & 81.8 & 83.5  &  94.1   &   86.4   \\
DeepLab         &  76.2  &  72.8  &  94.6  &  81.2      \\
\hline
Ours         & \textbf{83.9}  &  \textbf{83.6}   & \textbf{94.1} &      \textbf{87.2}     \\
\hline
\end{tabular}\vspace{-0.5em}
\caption{Comparison of geometric surface labeling performance with three state-of-the-art methods over LM+SUN dataset.}\vspace{-1em}
\label{table:LM+SUN}
\end{center}
\end{table}

\begin{table}\small
\renewcommand{\arraystretch}{1.1}
\addtolength{\tabcolsep}{-1pt}
\begin{center}
\begin{tabular}{c|cc}
\hline
Method & Subclasses & Main classes \\
\hline
Hoiem \textit{et al.}  & 68.8 & 89.0  \\
Superparsing & 73.7 & 88.2  \\
Liu \textit{et al.} & 76.3 & -  \\
\hline
Ours         & \textbf{80.1}  & \textbf{91.8}         \\
\hline
\end{tabular}\vspace{-0.5em}
\caption{Comparison of geometric surface labeling performance with three state-of-the-arts methods in terms of mean accuracy on Geometric Context dataset.}\vspace{-0.5em}
\label{table:GeometricContext}
\end{center}
\end{table}

\noindent
\textbf{Interaction Relation Prediction.} The MS-LSTM sub-network can predict the interaction relation results for two adjacent super-pixels. Note that we use five MS-LSTM layers and five scales of super-pixel maps are sequentially employed, including 128, 64, 48, 32, 16 super-pixels in five layers. The H-LSTM outputs the  interaction relation prediction results after each MS-LSTM layer to enable the  deep supervision for better feature  learning. Table \ref{table:CorrelationAnalysis} shows the average precision after passing different number of MS-LSTM layers. The improvements can be  observed on most of datasets by gradually using more MS-LSTM layers. It verifies  well the effectiveness of exploiting more discriminative feature  representation based on the hierarchical multi-scale super-pixel LSTM. The hierarchical MS-LSTM enables the model to simultaneously capture the global geometric structure information by increasingly sensing the larger contextual region and also keep track of local fine details by remembering the local interaction of small super-pixels.

\begin{table}\small
\renewcommand{\arraystretch}{1.1}
\addtolength{\tabcolsep}{-0.5pt}
\begin{center}
\begin{tabular}{c|cccc}
\hline
\tabincell{c}{The number of\\ MS-LSTM layers}
                & SIFT-Flow & LM+SUN & G-Context \\
\hline
H-LSTM\_1   & 85.8 & 89.1 & 87.8 \\
H-LSTM\_2   & 89.8 & 94.7 & 90.6 \\
H-LSTM\_3    & 90.3 & 95.6 & 89.8 \\
H-LSTM\_4    & 90.4 & \textbf{96.7} & 90.7 \\
H-LSTM    & \textbf{91.2} & 95.8 & \textbf{90.8} \\
\hline
\end{tabular}\vspace{-0.5em}
   \caption{Comparisons of interaction relation prediction results (Average Precision) by using different number of MS-LSTM layers on three datasets. ``H-LSTM\_1", ``H-LSTM\_2", ``H-LSTM\_3", ``H-LSTM\_4" represent the results using 1,2,3,4 MS-LSTM layers, respectively.}\vspace{-1.3em}
   \label{table:CorrelationAnalysis}
\end{center}
\end{table}

\vspace{-0.5em}
\subsection{Ablative Study}

We further evaluate different architecture variants to verify the effectiveness of the important components in our H-LSTM, presented in Table~\ref{table:ComponentAnalysis}.

\noindent
\textbf{Comparison with convolutional layers.} To strictly evaluate the effectiveness of using the proposed P-LSTM layer, we report the performance of purely using convolutional layers, i.e. ``convolution". To make fair comparison with P-LSTM layer, we utilize five convolutional layers, each of which contains $576 = 64 \times 9$ convolutional filters with size $3 \times 3$, because nine LSTMs are used in a P-LSTM layer and each of them has 64 hidden cell outputs. Compared with ``H-LSTM (ours)", ``convolution" decreases the pixel accuracy. It demonstrates the superiority of using P-LSTM layers to harness complex long-distances dependencies over convolutional layers.

\noindent
\textbf{Multi-task learning.} Note that we jointly optimize the geometric surface labeling and relation prediction task within a unified network. We demonstrate the effectiveness of multi-task learning by comparing our H-LSTM  with the version that only predicts the geometric surface labeling, i.e. ``P-LSTM". The supervision information for interaction relation and MS-LSTM networks are discarded in  ``P-LSTM". The large performance decrease speaks well that these two tasks can mutually benefit from each other and help learn more  meaningful and discriminative features.

\noindent
\textbf{Comparison with single scale of super-pixel map.} We also validate the advantage of using multi-scale super-pixel representation in the MS-LSTM sub-network on interaction relation prediction. ``S-LSTM" shows the results of using the same scale of super-pixels (i.e. 48 super-pixels) in each S-LSTM layer. The improvement of ``H-LSTM" over ``P-LSTM+S-LSTM" demonstrates that the richer contextual dependencies can be captured by using hierarchical multi-scale feature learning.

\begin{table}\small
\renewcommand{\arraystretch}{1.1}
\addtolength{\tabcolsep}{-1pt}
\begin{center}
\begin{tabular}{c|ccccc|ccccc}
\hline
Model settings & SIFT-Flow & LM+SUN\\
\hline
Convolution       & 94.66  & 89.92\\
P-LSTM            & 94.68 & 90.13\\
P-LSTM + S-LSTM   & 95.24 & 91.06 \\
H-LSTM (ours)  & \textbf{95.41}  & \textbf{91.34} \\
\hline
\end{tabular}\vspace{-0.5em}
\caption{Performance comparisons with different variants of our method in terms of pixel accuracy.}\vspace{-0.5em}
\label{table:ComponentAnalysis}
\end{center}
\end{table}

\vspace{-0.5em}
\subsection{Application to 3D Reconstruction}

\begin{figure}[t]
\centering
\includegraphics[width= 2.8 in]{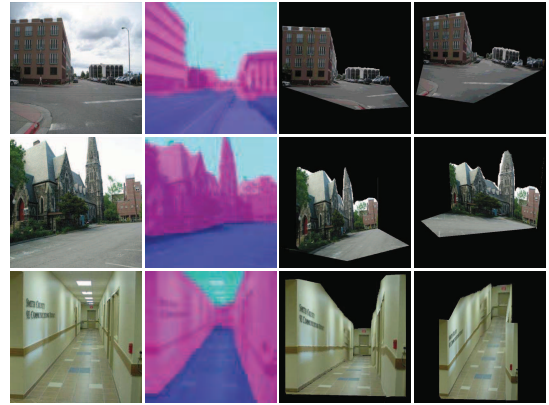}\vspace{-1em}
   \caption{Some results of single-view 3D reconstruction. The first column is the original image. The second column is the geometric surface labeling result and the last two columns are the reconstruction results from two different views. }\vspace{-1em}
\label{fig:3D-Reconstruction}
\end{figure}

Our main geometric class labels and interaction relation prediction over regions are sufficient to reconstruct scaled 3D models of many scenes. Figure \ref{fig:3D-Reconstruction} shows some scene images and  the reconstructed 3D scenes generated based on our geometric parsing results. Besides the obvious graphic applications, e.g. creating virtual walkthroughs, we believe that extra valuable information could be provided by such models to other artificial intelligence applications.

\vspace{-0.5em}
\section{Conclusion}

In this paper, we have proposed a multi-scale and context-aware scene paring model via recurrent Long Short-Term Memory neural network. Our approach have demonstrated a new state-of-the-art on the problem of geometric scene parsing, and also impressive results on 3D reconstruction from still images.

%% The file named.bst is a bibliography style file for BibTeX 0.99c
\bibliographystyle{named}
\bibliography{ijcai16}

\begin{thebibliography}{}

\bibitem[\protect\citeauthoryear{Boyd and Vandenberghe}{2004}]{Book:LSE}
Stephen Boyd and Lieven Vandenberghe.
\newblock {\em Convex optimization}.
\newblock Cambridge university press, 2004.

\bibitem[\protect\citeauthoryear{Byeon \bgroup \em et al.\egroup
  }{2014}]{Conf:2DLSTM}
Wonmin Byeon, Marcus Liwicki, and Thomas~M Breuel.
\newblock Texture classification using 2d lstm networks.
\newblock In {\em ICPR}, 2014.

\bibitem[\protect\citeauthoryear{Byeon \bgroup \em et al.\egroup
  }{2015}]{Conf:LSTM-Labeling}
Wonmin Byeon, Thomas~M. Breuel, Federico Raue, and Marcus Liwicki.
\newblock Scene labeling with {LSTM} recurrent neural networks.
\newblock In {\em CVPR}, 2015.

\bibitem[\protect\citeauthoryear{Chen \bgroup \em et al.\egroup
  }{2015}]{Conf:DeepLab}
Liang-Chieh Chen, George Papandreou, Iasonas Kokkinos, Kevin Murphy, and Alan~L
  Yuille.
\newblock Semantic image segmentation with deep convolutional nets and fully
  connected crfs.
\newblock In {\em ICLR}, 2015.

\bibitem[\protect\citeauthoryear{Graves \bgroup \em et al.\egroup
  }{2007}]{Conf:MultidimensionalRNN}
A.~Graves, S.~Fernandez, and J.~Schmidhuber.
\newblock Multidimensional recurrent neural networks.
\newblock In {\em ICANN}, 2007.

\bibitem[\protect\citeauthoryear{Graves \bgroup \em et al.\egroup
  }{2013}]{Conf:Speech-RNN}
A.~Graves, A.~Mohamed, and G.~Hinton.
\newblock Speech recognition with deep recurrent neural networks.
\newblock In {\em ICASSP}, 2013.

\bibitem[\protect\citeauthoryear{Hochreiter and Schmidhuber}{1997}]{LSTM}
Sepp Hochreiter and J{\"u}rgen Schmidhuber.
\newblock Long short-term memory.
\newblock {\em Neural computation}, 9(8):1735--1780, 1997.

\bibitem[\protect\citeauthoryear{Hoiem \bgroup \em et al.\egroup
  }{2005}]{DBLP:Reconstruction}
Derek Hoiem, Alexei~A. Efros, and Martial Hebert.
\newblock Automatic photo pop-up.
\newblock {\em {ACM} Trans. Graph.}, 24(3):577--584, 2005.

\bibitem[\protect\citeauthoryear{Hoiem \bgroup \em et al.\egroup
  }{2007}]{hoiem2007recovering}
Derek Hoiem, Alexei~A Efros, and Martial Hebert.
\newblock Recovering surface layout from an image.
\newblock {\em International Journal of Computer Vision}, 75(1):151--172, 2007.

\bibitem[\protect\citeauthoryear{Hoiem \bgroup \em et al.\egroup
  }{2008}]{hoiem2008closing}
Derek Hoiem, Alexei Efros, Martial Hebert, et~al.
\newblock Closing the loop in scene interpretation.
\newblock In {\em Computer Vision and Pattern Recognition, 2008. CVPR 2008.
  IEEE Conference on}, pages 1--8. IEEE, 2008.

\bibitem[\protect\citeauthoryear{Kalchbrenner \bgroup \em et al.\egroup
  }{2015}]{Arxiv:GridLSTM}
Nal Kalchbrenner, Ivo Danihelka, and Alex Graves.
\newblock Grid long short-term memory.
\newblock {\em arXiv preprint arXiv:1507.01526}, 2015.

\bibitem[\protect\citeauthoryear{Kanji}{2015}]{Conf:Robots}
Tanaka Kanji.
\newblock Unsupervised part-based scene modeling for visual robot localization.
\newblock In {\em ICRA}, 2015.

\bibitem[\protect\citeauthoryear{Kr{\"{a}}henb{\"{u}}hl and
  Koltun}{2011}]{Conf:CRF1}
Philipp Kr{\"{a}}henb{\"{u}}hl and Vladlen Koltun.
\newblock Efficient inference in fully connected crfs with gaussian edge
  potentials.
\newblock In {\em NIPS}, 2011.

\bibitem[\protect\citeauthoryear{Ladicky \bgroup \em et al.\egroup
  }{2009}]{Conf:CRF2}
Lubor Ladicky, Christopher Russell, Pushmeet Kohli, and Philip H.~S. Torr.
\newblock Associative hierarchical crfs for object class image segmentation.
\newblock In {\em ICCV}, 2009.

\bibitem[\protect\citeauthoryear{Liang \bgroup \em et al.\egroup
  }{2015}]{Conf:LG-LSTM}
Xiaodan Liang, Xiaohui Shen, Donglai Xiang, Jiashi Feng, Liang Lin, and
  Shuicheng Yan.
\newblock Semantic object parsing with local-global long short-term memory.
\newblock {\em arXiv preprint arXiv:1511.04510}, 2015.

\bibitem[\protect\citeauthoryear{Liu \bgroup \em et al.\egroup
  }{2011a}]{DBLP:SiftFlow}
Ce~Liu, Jenny Yuen, and Antonio Torralba.
\newblock Nonparametric scene parsing via label transfer.
\newblock {\em {IEEE} Trans. Pattern Anal. Mach. Intell.}, 33(12):2368--2382,
  2011.

\bibitem[\protect\citeauthoryear{Liu \bgroup \em et al.\egroup
  }{2011b}]{Conf:SuperPixelGeneration}
Ming-Yu Liu, Oncel Tuzel, Srikumar Ramalingam, and Rama Chellappa.
\newblock Entropy rate superpixel segmentation.
\newblock In {\em CVPR}, 2011.

\bibitem[\protect\citeauthoryear{Liu \bgroup \em et al.\egroup
  }{2014}]{Conf:SingleView}
Xiaobai Liu, Yibiao Zhao, and Song{-}Chun Zhu.
\newblock Single-view 3d scene parsing by attributed grammar.
\newblock In {\em CVPR}, 2014.

\bibitem[\protect\citeauthoryear{Liu \bgroup \em et al.\egroup
  }{2015}]{Conf:CNN-MRF}
Ziwei Liu, Xiaoxiao Li, Ping Luo, Chen~Change Loy, and Xiaoou Tang.
\newblock Semantic image segmentation via deep parsing network.
\newblock In {\em ICCV}, 2015.

\bibitem[\protect\citeauthoryear{Long \bgroup \em et al.\egroup
  }{2015}]{Conf:FCN}
Jonathan Long, Evan Shelhamer, and Trevor Darrell.
\newblock Fully convolutional networks for semantic segmentation.
\newblock In {\em CVPR}, 2015.

\bibitem[\protect\citeauthoryear{Mobahi \bgroup \em et al.\egroup
  }{2011}]{Conf:Holistic3D}
Hossein Mobahi, Zihan Zhou, Allen~Y. Yang, and Yi~Ma.
\newblock Holistic 3d reconstruction of urban structures from low-rank
  textures.
\newblock In {\em ICCV Workshops}, 2011.

\bibitem[\protect\citeauthoryear{Nieuwenhuisen \bgroup \em et al.\egroup
  }{2010}]{Conf:IndoorNavigation}
Matthias Nieuwenhuisen, J{\"o}rg St{\"u}ckler, and Sven Behnke.
\newblock Improving indoor navigation of autonomous robots by an explicit
  representation of doors.
\newblock In {\em ICRA}, 2010.

\bibitem[\protect\citeauthoryear{Pinheiro and
  Collobert}{2015}]{Conf:ImageToPixelLabel}
Pedro H.~O. Pinheiro and Ronan Collobert.
\newblock From image-level to pixel-level labeling with convolutional networks.
\newblock In {\em CVPR}, 2015.

\bibitem[\protect\citeauthoryear{Russell \bgroup \em et al.\egroup
  }{2008}]{DBLP:LabelMe}
Bryan~C. Russell, Antonio Torralba, Kevin~P. Murphy, and William~T. Freeman.
\newblock Labelme: {A} database and web-based tool for image annotation.
\newblock {\em International Journal of Computer Vision}, 77(1-3):157--173,
  2008.

\bibitem[\protect\citeauthoryear{Saxena \bgroup \em et al.\egroup
  }{2009}]{DBLP:Make3D}
Ashutosh Saxena, Min Sun, and Andrew~Y. Ng.
\newblock Make3d: Learning 3d scene structure from a single still image.
\newblock {\em {IEEE} Trans. Pattern Anal. Mach. Intell.}, 31(5):824--840,
  2009.

\bibitem[\protect\citeauthoryear{Simonyan and Zisserman}{2015}]{Conf:VGG-16}
Karen Simonyan and Andrew Zisserman.
\newblock Very deep convolutional networks for large-scale image recognition.
\newblock In {\em ICLR}, 2015.

\bibitem[\protect\citeauthoryear{Socher \bgroup \em et al.\egroup
  }{2011}]{Conf:Recursive_Socher}
Richard Socher, Cliff~Chiung{-}Yu Lin, Andrew~Y. Ng, and Christopher~D.
  Manning.
\newblock Parsing natural scenes and natural language with recursive neural
  networks.
\newblock In {\em ICML}, 2011.

\bibitem[\protect\citeauthoryear{Tighe and Lazebnik}{2013}]{DBLP:LM+SUN}
Joseph Tighe and Svetlana Lazebnik.
\newblock Superparsing - scalable nonparametric image parsing with superpixels.
\newblock {\em International Journal of Computer Vision}, 101(2):329--349,
  2013.

\bibitem[\protect\citeauthoryear{Xiao \bgroup \em et al.\egroup
  }{2010}]{Conf:SunDataset}
Jianxiong Xiao, James Hays, Krista~A. Ehinger, Aude Oliva, and Antonio
  Torralba.
\newblock {SUN} database: Large-scale scene recognition from abbey to zoo.
\newblock In {\em CVPR}, 2010.

\end{thebibliography}

\end{document}